\documentclass{article}

\usepackage[preprint,nonatbib]{neurips_2024}

\usepackage[utf8]{inputenc} % allow utf-8 input
\usepackage[T1]{fontenc}    % use 8-bit T1 fonts
\usepackage{hyperref}       % hyperlinks
\usepackage{url}            % simple URL typesetting
\usepackage{booktabs}       % professional-quality tables
\usepackage{amsfonts}       % blackboard math symbols
\usepackage{nicefrac}       % compact symbols for 1/2, etc.
\usepackage{microtype}      % microtypography
\usepackage{xcolor}         % colors

\usepackage{color,xcolor}
\usepackage{amsmath}
\usepackage{graphicx}
\usepackage{booktabs}
\usepackage{multirow}
\usepackage{soul}
\usepackage{float}
\usepackage{footmisc}
\DeclareMathOperator*{\argmin}{arg\,min}
\definecolor{mylightgreen}{RGB}{144,238,144}

\title{Time Cell Inspired Temporal Codebook in Spiking Neural Networks for Enhanced Image Generation}

\author{%
  {Linghao Feng$^{1,4*}$, Dongcheng Zhao$^{1,2*}$, Sicheng Shen$^{1,4}$} \\
  \textbf{Yiting Dong$^{1,2,4}$, Guobin Shen$^{1,2,4}$, Yi Zeng$^{1,2,3,4,5\dagger}$}\\
  $^1$Brain-inspired Cognitive Intelligence Lab, Institute of Automation, Chinese Academy of Sciences\\  
    $^2$Center for Long-term Artificial Intelligence\\
  $^3$Key Laboratory of Brain Cognition and Brain-inspired Intelligence Technology, CAS \\
  $^4$School of Future Technology, University of Chinese Academy of Sciences \\
   $^5$School of Artificial Intelligence, University of Chinese Academy of Sciences \\
  \texttt{\{fenglinghao2022, zhaodongcheng2016,shensicheng2024,} \\
  \texttt{dongyiting2020,shenguobin2021,yi.zeng@ia.ac.cn\}} \\
}

\begin{document}

\maketitle

\renewcommand{\thefootnote}{\fnsymbol{footnote}}
\footnotetext[2]{Corresponding Author.}
\footnotetext[1]{Equal contribution.}

\begin{abstract}
This paper presents a novel approach leveraging Spiking Neural Networks (SNNs) to construct a Variational Quantized Autoencoder (VQ-VAE) with a temporal codebook inspired by hippocampal time cells. This design captures and utilizes temporal dependencies, significantly enhancing the generative capabilities of SNNs. Neuroscientific research has identified hippocampal "time cells" that fire sequentially during temporally structured experiences. Our temporal codebook emulates this behavior by triggering the activation of time cell populations based on similarity measures as input stimuli pass through it. We conducted extensive experiments on standard benchmark datasets, including MNIST, FashionMNIST, CIFAR10, CelebA, and downsampled LSUN Bedroom, to validate our model's performance. Furthermore, we evaluated the effectiveness of the temporal codebook on neuromorphic datasets NMNIST and DVS-CIFAR10, and demonstrated the model's capability with high-resolution datasets such as CelebA-HQ, LSUN Bedroom, and LSUN Church. The experimental results indicate that our method consistently outperforms existing SNN-based generative models across multiple datasets, achieving state-of-the-art performance. Notably, our approach excels in generating high-resolution and temporally consistent data, underscoring the crucial role of temporal information in SNN-based generative modeling.
\end{abstract}

%%%%%%%%%%%%%%%%%%%%%%%%%%%%%%%%%%%%%%%%%%%%%% Introduction %%%%%%%%%%%%%%%%%%%%%%%%%%%%%%%%%%%%%%%%%%%%%%%%%%%%%%%%%%%%
\section{Introduction}
Episodic memory~\cite{hassabis2007deconstructing} is considered to be constructive; the act of recalling is conceptualized as the construction of a past experience rather than the retrieval of an exact replica~\cite{bartlett1995remembering,schacter2012constructive}. Generative models such as the variational autoencoder (VAE)~\cite{kingma2013auto} and diffusion model~\cite{ho2020denoising} aim to generate new data that closely resembles the original dataset. This involves sampling from an approximate distribution of real data to create a probabilistic model that represents the real world. By correlating the constructive memory process of biological organisms with the sampling processes employed by generative models, a natural parallel can be drawn between the functionalities of generative models in machine learning and the construction of episodic memory in biological organisms.

In the field of neuroscience, the hippocampus plays a crucial role in memory. It facilitates memory consolidation~\cite{marshall2007contribution}, supports spatial navigation~\cite{eichenbaum2017role}, and is integral to the formation and recall of episodic memories~\cite{burgess2002human}. It is pivotal in both spatial~\cite{o1976place} and non-spatial~\cite{squire1992memory} memory processes.  Several studies~\cite{manns2007gradual,pastalkova2008internally,macdonald2011hippocampal,kraus2013hippocampal} have revealed that in addition to place cells~\cite{best2001spatial}, which fire when rats are in a particular location in a spatially structured environment, the hippocampus contains time cells~\cite{eichenbaum2014time}, which fire at particular moments in a temporally structured period. The discovery of time cells highlights temporal coding as a stable and prevalent feature of hippocampal firing patterns.% This finding provides new perspectives on understanding the role of the hippocampus in processing temporal information and lays the groundwork for further studies on the complex functions of the hippocampus in memory mechanisms.

Architectures akin to autoencoders~\cite{zhai2018autoencoder}, featuring encoder and decoder structures, are commonly employed for hippocampal modeling~\cite{whittington2020tolman,nagy2020optimal,van2020brain}. The process of feature extraction from input stimuli by the encoder is analogous to the inferential function of the hippocampus~\cite{sanders2020hippocampal}, where real-world inputs are transformed into more abstract representations. Moreover, the decoder reconstructs the actual external stimuli from these abstract features, akin to the replay function of the hippocampus~\cite{olafsdottir2018role}. VAEs have been used as models for hippocampal memory consolidation~\cite{spens2024generative}. Similarly, Helmholtz machines have been employed to model hippocampal function for path integration~\cite{george2024generative}. However, despite previous studies on hippocampal generative models effectively modeling certain aspects of hippocampal functions, computational studies on the encoding of temporal information by time cells remain insufficient. 

In this work, we utilize the Vector Quantized-Variational AutoEncoder (VQ-VAE) due to its robust data compression and reconstruction capabilities. The VQ-VAE demonstrates superior performance compared to the vanilla VAE, particularly due to its discrete latent space, which enhances interpretability. A significant challenge in hippocampus-like generative models is the integration of temporal information conveyed by time cells. To address this, we employ Spiking Neural Networks (SNNs) to construct the Spiking VQ-VAE. Unlike artificial neural networks (ANNs), SNNs operate more similarly to neurons in the real brain. They convey information through spikes rather than floating-point numbers, with inputs and outputs typically consisting of temporally structured spike sequences. Leveraging these temporal properties of SNNs, we integrate temporal information into the VQ-VAE codebook by introducing a "temporal codebook," specifically designed to capture and utilize time-dependent information, simulating the function of time cells in the hippocampus. This design not only emulates hippocampal mechanisms in processing episodic memory but also enhances generative capabilities across multiple datasets. In summary, the contributions of this study are:
\begin{itemize}
  \item We propose the Spiking VQ-VAE with a temporal codebook that integrates the temporal characteristics of hippocampal time cells, capturing and leveraging time-dependent information to enhance generative capabilities.
  \item Through experimental validation across multiple datasets, we demonstrate the effectiveness of the proposed method, achieving superior performance compared to other SNN-based generative models.
  \item Compared to previous models, our approach generates higher-resolution stimuli and more temporally coherent neuromorphic data, showcasing significant improvements in high-resolution and temporal data generation.
\end{itemize}

%%%%%%%%%%%%%%%%%%%%%%%%%%%%%%%%%%%%%%%%%%%%%%% Related Work %%%%%%%%%%%%%%%%%%%%%%%%%%%%%%%%%%%%%%%%%%%%%%%%%%%%%%%%%%
\section{Related Work}
\paragraph{Variational Autoencoder}
The Variational Autoencoder (VAE)\cite{kingma2013auto} is a generative model that imposes constraints on latent variables within the Autoencoder (AE) framework. Building on VAE, subsequent research introduced beta-VAE\cite{higgins2017beta}, which facilitates feature disentanglement in the latent space. To enhance the information content of latent variables, InfoVAE~\cite{zhao2017infovae} was developed, which leverages mutual information to increase the correlation between latent variables and input stimuli. For multimodal data, researchers modified the VAE's Evidence Lower Bound (ELBO), resulting in Multimodal VAEs~\cite{khattar2019mvae,wu2018multimodal}. The Hierarchical VAE (HVAE)\cite{vahdat2020nvae,child2020very} was introduced to incorporate multiple latent variables, creating a hierarchical structure within the latent space. More recently, to achieve discrete representations in the VAE latent space, VQ-VAE\cite{van2017neural} was proposed, depicting the latent space by learning a discrete codebook. Building on VQ-VAE, VQ-GAN~\cite{esser2021taming} further improved performance by replacing the mean squared error (MSE) loss with a discriminator, utilizing adversarial learning.

\paragraph{Generative Models Based on SNNs}
Several studies have been designed to explore the potential of SNNs in generative tasks.~\cite{kamata2022fully} introduced a Fully Spiking Variational Autoencoder (FSVAE), which samples images according to the Bernoulli distribution.~\cite{kotariya2022spiking} developed a fully SNN-based backbone with a time-to-first-spike coding scheme, while~\cite{feng2023sgad} constructed a spiking generative adversarial network with attention-scoring decoding to address temporal inconsistency issues.~\cite{liu2023spiking} proposed a spike-based Vector Quantized Variational Autoencoder (VQ-SVAE) to learn a discrete latent space for images, and~\cite{cao2024spiking} introduced a spiking version of the Denoising Diffusion Probabilistic Model (DDPM) with a threshold-guided strategy. Although numerous SNN-based generative models have been proposed, their generative capabilities remain relatively weak, making it challenging to generate high-quality and temporally coherent data.
\section{Methods}
% In this section, we introduce the method of Spiking VQVAE with a Temporal Codebook, enabling the generation of higher-quality and more temporally coherent images. The subsections cover the Leaky Integrate-and-Fire (LIF) model, an introduction to the Temporal Codebook, autoregressive image generation based on Spiking Transformers, and training losses for perceptual quality enhancement.

% \subsection{Spiking VQVAE Framework}

\begin{figure}[t]
  \centering
  \includegraphics[width=0.95\textwidth]{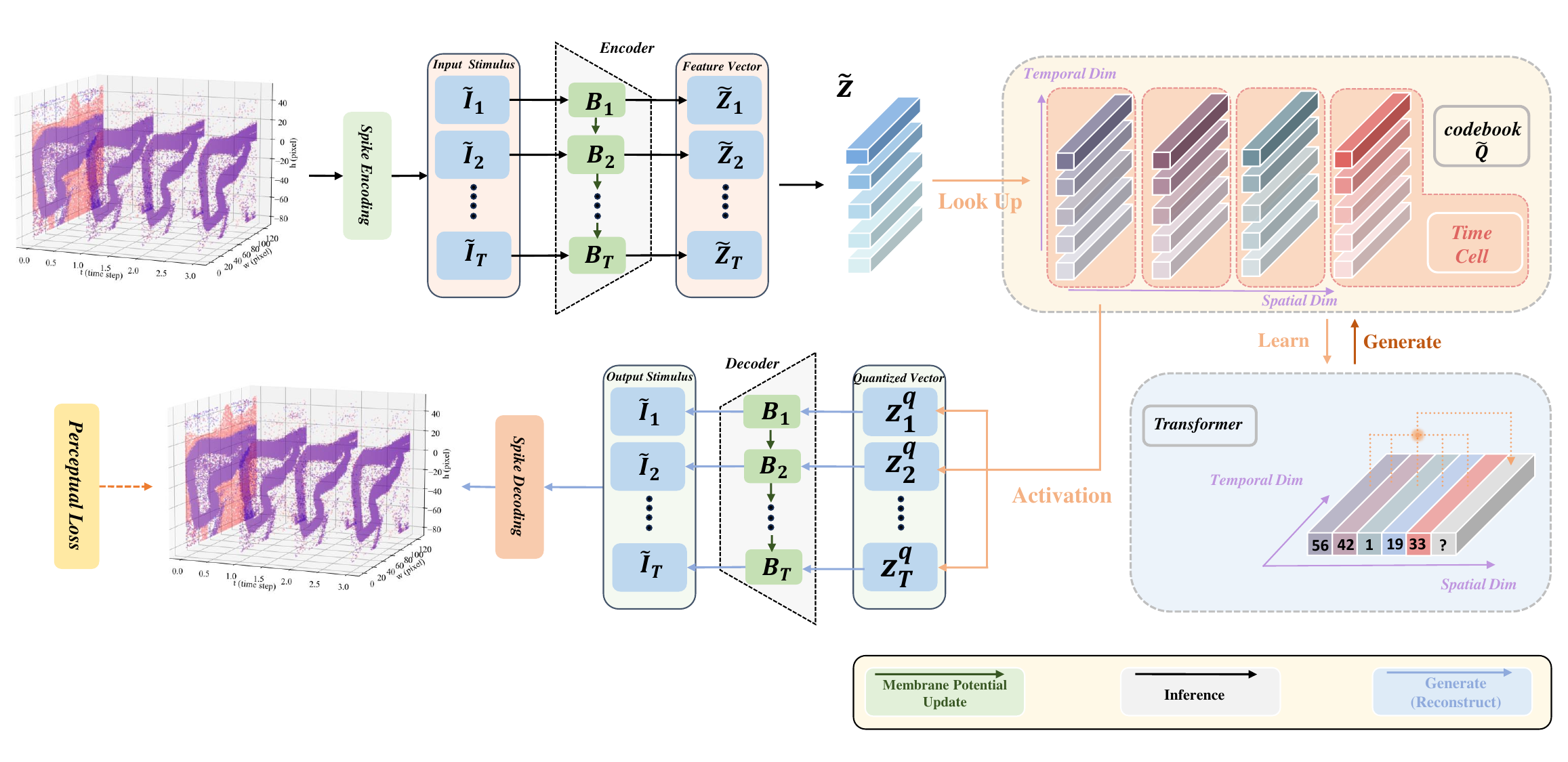}
  \caption{The overview of the proposed method. After spike encoding, the input stimulus \( \tilde{I}_t \), is processed by the encoder to yield \( z_t \). Concatenating \( z_t \) and passing it through the temporal codebook \( \tilde{Q} \) produces quantized \( z^q_t \), which is then input to the decoder to reconstruct the stimulus \(X_t\).}
  \label{fig:overview}
\end{figure}

The overall training pipeline of Spiking VQVAE is shown in Figure~\ref{fig:overview}. First, we convert the input  \(I \in \mathbb{R}^{C \times H \times W}\)  into a spike form using direct coding, resulting in \(\tilde{I} = SE(I) \in \mathbb{R}^{T \times C \times H \times W}\), where \(SE(\cdot)\) denotes the spike encoder and \(T\) represents the total time steps. Next, we use an encoder, denoted as \(E(\cdot)\), to extract features from \(\tilde{I}\). At each time step \(t\), the feature extraction is represented as \(z_t = E(\tilde{I}_t) \in \mathbb{R}^{c \times h \times w}\), where \(c\), \(h\), and \(w\) are the dimensions of the extracted features.

By concatenating the feature vectors \(z_t\) from each time step, we construct a temporally-informed feature vector \(\tilde{z} = [z_1, \ldots, z_T] \in \mathbb{R}^{T \times c \times h \times w}\) and then quantize it using the Temporal Codebook \(\tilde{Q}(\cdot)\), resulting in (\(\tilde{z^q} = \tilde{Q}(\tilde{z})\)). The quantized features are then input into a decoder, denoted as \(D(\cdot)\). At each time step \(t\), the decoding process is represented as \(x_t = D(\tilde{z^{q}_{t}})\), where \(x_t\) is the generated stimulus. Finally, by passing the decoded features from each time step through a spike decoder \(SD(\cdot)\), we obtain a static feature representation \(X = SD(x_1, \ldots, x_T) \in \mathbb{R}^{C \times H \times W}\).

The architecture of the encoder \(E\) and decoder \(D\) employs a spiking convolutional neural network and a spiking Transformer framework~\cite{li2022spikeformer}.

\subsection{Leaky Integrate-and-Fire Model}

We utilize the Leaky Integrate-and-Fire (LIF) neuron model~\cite{burkitt2006review} due to its simplicity and effectiveness in emulating the dynamic behavior of biological neurons. The discretized form of the LIF neuron dynamics is:
\begin{align}
   & H[t] = V[t-1] + \frac{1}{\tau}(I[t] - (V[t-1] - V_{reset})) \notag\\
   & S[t] = \Theta(H[t] - V_{th}) \label{eq:lif_discretized} \\
   & V[t] = H[t](1 - S[t]) + S[t] V_{reset} \notag
\end{align}
Here, \(\tau\) is the membrane time constant, \(I[t]\) is the input synaptic current, \(H[t]\) is the membrane potential after charging but before firing a spike, and \(S[t]\) represents the spike generated when the membrane potential exceeds the threshold \(V_{th}\), resetting the potential to \(V_{reset}\). The Heaviside step function \(\Theta(v)\) is 1 when \(v \geq 0\) and 0 otherwise. Since \(\Theta\) is non-differentiable, we have employed the surrogate gradient method~\cite{neftci2019surrogate} to optimize the SNN network.

\subsection{Time Cell Inspired Temporal Codebook}

% In VQ-VAE, the codebook is a critical component that bridges the encoder and decoder. The codebook consists of a set of discrete embedding vectors, each representing a unique quantized latent state. During the encoding process, the continuous latent representation generated by the encoder is mapped to the nearest embedding vector in the codebook, typically based on Euclidean distance. Specifically, it employs a static codebook \(Q = \{q_k\}_{k=1}^{N_q}\) (we use the symbol \(Q\) to denote the vanilla codebook, distinguishing it from the temporal codebook \(\tilde{Q}\)), where \(q_k \in \mathbb{R}^{c}\) and \(N_q\) is the number of quantization vectors. For an input \(\tilde{z}\), the vanilla method extracts each \(z_t\) from \(\tilde{z}\) and compares it with \(\{q_k\}\), selecting the quantized \(z^{q}_{t}\) according to the following minimum distance rule:

In VQ-VAE, the codebook is essential for linking the encoder and decoder. It consists of discrete embedding vectors, each representing a unique quantized latent state. During encoding, the continuous latent representation is mapped to the nearest embedding vector in the codebook, typically based on Euclidean distance. Specifically, a static codebook \(Q = \{q_k\}_{k=1}^{N_q}\) is employed, where \(q_k \in \mathbb{R}^{c}\) and \(N_q\) is the number of quantization vectors. For an input \(\tilde{z}\), each \(z_t\) is extracted and compared to \(\{q_k\}\), selecting the quantized \(z^{q}_{t}\) according to the minimum distance rule as shown in Eq.~\ref{eq:v_codebook}.

\begin{equation}
  [z^{q}_{t}]_{ij} = \left( \argmin_{q_k \in Q} ||[z_t]_{ij} - q_k|| \right) \in \mathbb{R}^{c}
  \label{eq:v_codebook}
\end{equation}
where \([\cdot]_{ij}\) denotes the vector at the \(i\)-th row and \(j\)-th column of the tensor \(z^q_t \in \mathbb{R}^{c \times h \times w}\). After obtaining \(z^q\), \(\tilde{z^q}\) is formed by concatenating \(z^q_t\).

This process, known as vector quantization, discretizes the latent space, allowing the model to learn compressed representations of input data. The decoder reconstructs data from these quantized latent states. During training, the codebook is updated to ensure the embedding vectors capture essential features of the input, enhancing VQ-VAE's generative capabilities. This mechanism enables VQ-VAE to sample from the discrete latent space to generate high-quality reconstructions and new samples.

While VQ-VAE shows promise in generative modeling, traditional implementations often struggle with temporal consistency in sequential data. The hippocampus, a critical brain region, encodes and retrieves temporal sequences, crucial for episodic memory formation. Its complex neural architecture supports a continuous temporal framework, integrating and recalling sequential experiences. Time cells, specialized neurons in the hippocampus, activate at specific moments within a sequence, encoding temporal aspects of experiences. These neurons are essential for maintaining the temporal structure of episodic memories, allowing the brain to distinguish between different time points and accurately reconstruct event sequences.

% This process, often referred to as vector quantization, effectively discretizes the latent space, enabling the model to learn compressed representations of the input data. The decoder then reconstructs the data from these quantized latent states. During training, the codebook is continuously updated to ensure the embedding vectors capture the essential features of the input data, thereby enhancing the generative capabilities of VQ-VAE. This mechanism allows VQ-VAE to sample from the discrete latent space defined by the codebook to generate high-quality reconstructions and new samples.

% While VQ-VAE demonstrates significant potential in various generative modeling applications, traditional implementations often struggle to maintain temporal consistency in sequential data. The hippocampus, a critical brain region, is involved in encoding and retrieving temporal sequences, playing a crucial role in the formation of episodic memory. Through its complex neural architecture, the hippocampus supports the creation of a continuous temporal framework, allowing sequential experiences to be seamlessly integrated and recalled. A key component of the hippocampal structure is time cells, a special class of neurons that activate at specific moments within a sequence, encoding the temporal aspect of experiences. These neurons are vital for maintaining the temporal structure of episodic memories, enabling the brain to distinguish between different time points and accurately reconstruct event sequences.

Inspired by the ability of hippocampal time cells to encode temporal information, we designed a temporal codebook \(\tilde{Q} = \{\tilde{q}_k\}_{k=1}^{N_q}\), where each \(\tilde{q}_k \in \mathbb{R}^{T \times c}\). Time cells provide a biological template for preserving temporal sequences, ensuring that events can be recalled in a coherent and temporally ordered manner. Our temporal codebook leverages this concept by incorporating time-sensitive embeddings to capture the dynamic changes in sequential data.

To retain temporal information in the quantized features, we use Equation \(\ref{eq:t_codebook}\) to calculate the quantized \(\tilde{z^q} \in \mathbb{R}^{T \times c \times h \times w}\):
\begin{equation}
  [\tilde{z^q}]_{ij} = \left( \argmin_{q^T_k \in Q} ||[\tilde{z^q}]_{ij} - \tilde{q}_k|| \right) \in \mathbb{R}^{T \times c}
  \label{eq:t_codebook}
\end{equation}

After obtaining \(\tilde{z^q}\), the loss function for updating the temporal codebook can be derived as follows:
\begin{equation}
  L_{Q^T} = ||sg(\tilde{z}) - \tilde{z^q}||^2_2 + \beta ||\tilde{z} - sg(\tilde{z^q})||^2_2
  \label{eq:loss_qt}
\end{equation}
where \(sg(\cdot)\) is the stop-gradient operator, and \(\beta\) is a hyperparameter for weighting.

\begin{figure}
  \centering
  \includegraphics[width=0.8\textwidth]{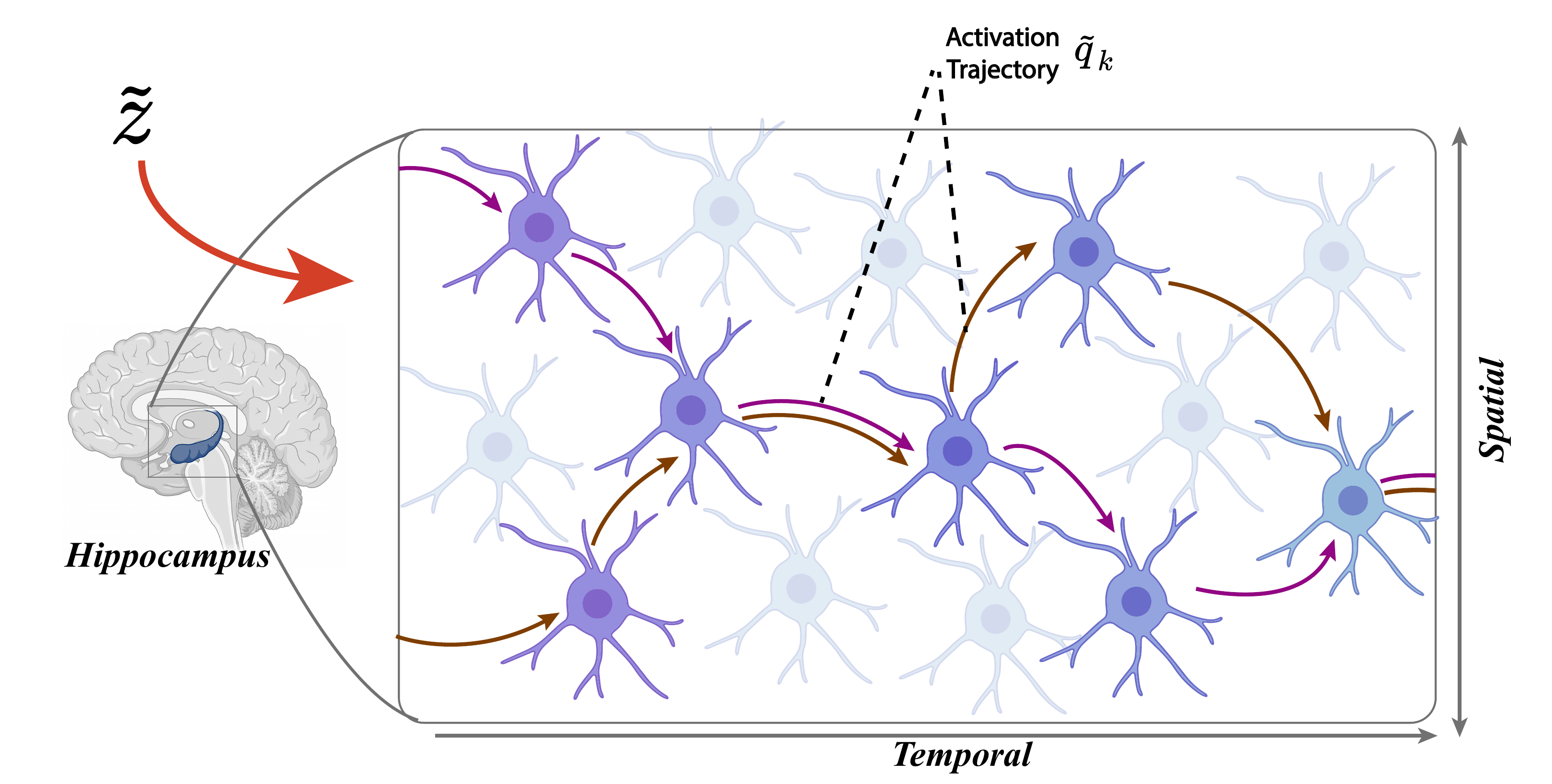}
  \caption{Spatiotemporal trajectories of neural activations from temporal codebook}
  \label{fig:trajectory}
\end{figure}

In \(\tilde{Q}\), each vector \(\tilde{q}_k \in \mathbb{R}^{T \times c}\) encompasses dimensions of temporal information \(T\) and spatial information \(h \times w\). Thus, this activation represents a spatiotemporal trajectory, akin to the activation of time cells, as illustrated in Figure~\ref{fig:trajectory}.

\subsection{Autoregressive Image Generation}

For the image generation component, we adopt the approach proposed by \cite{esser2021taming}, constructing an autoregressive language model using transformers \cite{vaswani2017attention}. We present both spiking and non-spiking versions of this model. The non-spiking version, based on a GPT-2-like structure with a 12-layer transformer, is employed for higher-quality image generation. The spiking version, incorporating LIF neurons and spiking self-attention mechanisms \cite{li2022spikeformer}, is designed for better temporal correspondence and greater energy efficiency. This model generates new images by producing sequences of indices from the temporal codebook. The sequence of indices \(s\) is modeled as:
\begin{equation}
  P(s) = \prod_{i} P(s_i | s_{<i})
  \label{eq:autoregressive}
\end{equation}
where \(s_i\) represents the index at position \(i\). The probability \(P(s_i | s_{<i})\) is approximated by the autoregressive model driven by transformers, as illustrated in Figure~\ref{fig:overview}. After obtaining the approximated index sequences \(s\), we retrieve the corresponding approximated \(\hat{\tilde{z_q}}\) from \(Q^T\). Subsequently, through the reconstruction process, we generate the stimulus \(\hat{X} = SD(D(\hat{\tilde{z_q}}))\).

\subsection{Perceptual Quality Enhancement Using Generative Adversarial Training}

To ensure high perceptual quality, we follow the methodology outlined in~\cite{esser2021taming}, utilizing perceptual loss~\cite{zhang2018unreasonable} and a generative adversarial training process to guide the model's loss and error propagation. Specifically, for the input stimulus \(I\) and the reconstructed stimulus \(X\), we extract features from both \(I\) and \(X\) using a pretrained VGG network. The perceptual loss is calculated based on the distance between the extracted features, as shown in Equation~\ref{eq:perceptual}:
\begin{equation}
  d(I,X) = \sum_{l} \frac{1}{H_l W_l} \sum_{h,w} ||w_l \odot ({f(I)}^{l}_{hw} - {f(X)}^{l}_{hw})||^2_2
  \label{eq:perceptual}
\end{equation}
where \({f(I)}^l \in \mathbb{R}^{H_l \times W_l \times C_l}\) and \({f(X)}^l \in \mathbb{R}^{H_l \times W_l \times C_l}\) are the features of \(I\) and \(X\) in the \(l\)-th layer of the pretrained VGG network. \(w_l\) is a learnable scale factor representing the importance of each layer. In practice, we incorporate perceptual loss, vanilla MSE loss, and discriminator loss as metrics for error propagation.

\section{Experiment}
\subsection{Setup}
To validate our method's efficacy, we conducted comprehensive experiments on various datasets. Standard benchmarks included MNIST, FashionMNIST, CIFAR10, CelebA, and downsampled LSUN Bedroom. For assessing the temporal codebook's effectiveness in temporally correlated generative tasks, we used neuromorphic datasets NMNIST~\cite{orchard2015converting} and DVS-CIFAR10~\cite{li2017cifar10}. Additionally, to demonstrate our method's capability on high-resolution datasets, we selected $256\times256$ resolution datasets CelebA-HQ~\cite{karras2017progressive}, LSUN Bedroom, and LSUN Church~\cite{yu15lsun}.
The experiments were implemented using the BrainCog framework~\cite{zeng2023braincog}. The base learning rate was set to \(4.5 \times 10^{-6}\) and adjusted according to batch size. AdamW was used as the optimizer, with the SNN time step varying from 2 to 6. The temporal codebook had \(N_q = 1024\) and each temporal step dimension \(c = 256\).
We evaluated our model's performance using Fréchet Inception Distance (FID)~\cite{heusel2017gans}, Inception Score (IS)~\cite{salimans2016improved}, Precision and Recall~\cite{sajjadi2018assessing}, and Kernel Inception Distance (KID)~\cite{binkowski2018demystifying}. In Sections 4.2 and 4.3, we utilized the second-stage autoregressive model based on SNN. In other parts of the experiment, we employed the second-stage autoregressive model based on ANN. Notably, for the first-stage VQ-VAE model, we consistently used the SNN-based model through all experiments.

\subsection{Comparative Analysis with SNN-based Generative Models}

\begin{figure}[h]
  \centering
  \includegraphics[width=0.95 \textwidth]{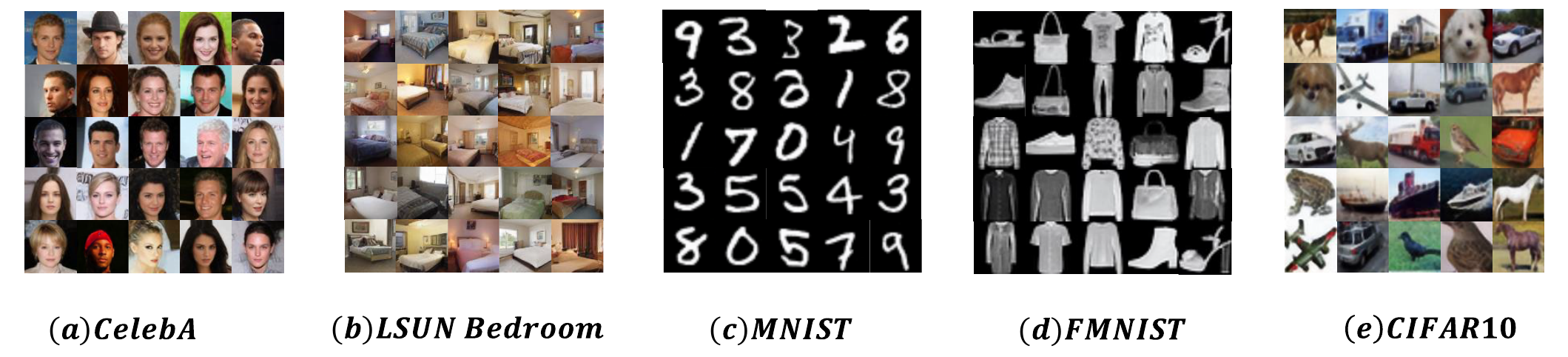}
  \caption{Comparative visualization of generated samples across various datasets}
  \label{fig:compare}
\end{figure}

We conducted an extensive evaluation of our proposed method against other SNN-based generative models using the CelebA, CIFAR10, MNIST, FashionMNIST, and downsampled LSUN Bedroom datasets. The visual results of our method are illustrated in Figure~\ref{fig:compare}. For quantitative assessment, we employed the FID across all datasets, and the IS for the CIFAR10 dataset.
The comparative results, detailed in Table~\ref{tab:compare}, demonstrate the superior performance of our model across various datasets. Notably, our model achieves superior results on all datasets except CIFAR10, where the FID score is marginally higher compared to the spiking diffusion models SDDPM and SDiT~\cite{cao2024spiking,yang2024sdit}. However, our model still surpasses these models in terms of the Inception Score on CIFAR10, underscoring the effectiveness of our approach.

% We conducte comparisons with other SNN-based generative models on the CelebA, CIFAR-10, MNIST, Fashion MNIST, and downsampled LSUN Bedroom datasets. The visual results of our method are shown in Figure~\ref{fig:compare}. The FID is measured across all these datasets. The Inception Score is assessed only for the CIFAR-10 dataset, which resembles ImageNet. For other datasets that do not closely resemble ImageNet, we do not measure Inception Score, as it makes little sense in this case.

% The comparative results are presented in Table~\ref{tab:compare}. The results indicate that, except for a slightly inferior FID on CIFAR-10 compared to Diffusion-based models~\cite{cao2024spiking,yang2024sdit}, our model outperforms other models on all other datasets. Additionally, our IS on CIFAR-10 is better than that of Diffusion-based models. Considering the inherent uncertainties in image generation metrics, this suggests that our results on CIFAR-10 are not lagging behind Diffusion-based models, but are instead comparable.

\begin{table}[h]
\resizebox{\linewidth}{!}{
\begin{tabular}{@{}cccccc@{}}
\toprule
\textbf{Dataset}                    & \textbf{Resolution}             & \textbf{Model}                          & \textbf{Time Steps} & \textbf{IS} $\uparrow$                      & \textbf{FID} $\downarrow$                    \\ \midrule
\multirow{5}{*}{CelebA$^{*}$}       & \multirow{5}{*}{$64 \times 64$} & FSVAE~\cite{kamata2022fully}            & 16                  & 3.70                                        & 101.60                                       \\
                                    &                                 & SGAD~\cite{feng2023sgad}                & 16                  & -                                           & 151.36                                       \\
                                    &                                 & SDDPM~\cite{cao2024spiking}             & 4                   & -                                           & 25.09                                        \\
                                    % &                                 & \textbf{Ours}                           & 4                   & -                                           & \sethlcolor{pink}\hl{\textbf{10.26}}         \\
                                    &                                 & \textbf{Ours}             & 4                   & -                                           & \textbf{13.11} \\ \midrule
\multirow{3}{*}{LSUN Bedroom$^{*}$} & \multirow{3}{*}{$64 \times 64$} & SDDPM~\cite{cao2024spiking}                                   & 4                   & -                                           & 47.64                                        \\
                                    % &                                 & \textbf{Ours}                           & 6                   & -                                           & \sethlcolor{pink}\hl{\textbf{8.83}}          \\
                                    &                                 & \textbf{Ours}             & 6                   & -                                           & \textbf{12.97} \\ \midrule
\multirow{7}{*}{MNSIT$^{*}$}        & \multirow{7}{*}{$28 \times 28$} & FSVAE~\cite{kamata2022fully}                                   & 16                  & 6.21                                        & 97.06                                        \\
                                    &                                 & SGAD~\cite{feng2023sgad}                                    & 16                  & -                                           & 69.64                                        \\
                                    &                                 & Spiking-Diffusion~\cite{liu2023spiking} & 16                  & -                                           & 37.50                                        \\
                                    &                                 & SDDPM~\cite{cao2024spiking}                                   & 4                   & -                                           & 29.48                                        \\
                                    &                                 & SDiT~\cite{yang2024sdit}                & 4                   & 2.45                                        & 5.54                                         \\
                                    % &                                 & \textbf{Ours}                           & 4                   & -                                           & \sethlcolor{pink}\hl{\textbf{4.46}}          \\
                                    &                                 & \textbf{Ours}             & 4                   & -                                           & \textbf{4.60}  \\ \midrule
\multirow{7}{*}{FashionMNIST$^{*}$} & \multirow{7}{*}{$28 \times 28$} & FSVAE~\cite{kamata2022fully}                                   & 16                  & 4.55                                        & 90.12                                        \\
                                    &                                 & SGAD~\cite{feng2023sgad}                                    & 16                  & -                                           & 165.42                                       \\
                                    &                                 & Spiking-Diffusion~\cite{liu2023spiking}                       & 16                  & -                                           & 91.98                                        \\
                                    &                                 & SDDPM~\cite{cao2024spiking}                                   & 4                   & -                                           & 21.38                                        \\
                                    &                                 & SDiT~\cite{yang2024sdit}                                    & 4                   & 4.55                                        & 5.49                                         \\
                                    % &                                 & \textbf{Ours}                           & 4                   & -                                           & \sethlcolor{pink}\hl{\textbf{3.41}}          \\
                                    &                                 & \textbf{Ours}             & 4                   & -                                           & \textbf{3.67}  \\ \midrule
\multirow{7}{*}{CIFAR-10}           & \multirow{7}{*}{$32 \times 32$} & FSVAE~\cite{kamata2022fully}                                   & 16                  & 2.95                                        & 175.50                                       \\
                                    &                                 & SGAD~\cite{feng2023sgad}                                    & 16                  & -                                           & 181.50                                       \\
                                    &                                 & Spiking-Diffusion~\cite{liu2023spiking}                       & 16                  & -                                           & 120.50                                       \\
                                    &                                 & SDDPM~\cite{cao2024spiking}                                   & 4                   & 7.66                                        & \textbf{16.98}         \\
                                    &                                 & SDiT~\cite{yang2024sdit}                                    & 4                   & 4.08                                        & 22.17 \\
                                    % &                                 & \textbf{Ours}                           & 4                   & \sethlcolor{pink}\hl{\textbf{8.99}}         & 30.30                                        \\
                                    &                                 & \textbf{Ours}             & 4                   & \textbf{8.86} & 31.24                                        \\ \bottomrule
\end{tabular}
}
\caption{Quantitative Comparison of Generative Models Across Multiple Datasets. ($^{*}$ denotes the omission of Inception Score (IS) measurements, as this dataset bears little resemblance to ImageNet, rendering the IS metrics relatively nonsensical in this context)}
\label{tab:compare}
\end{table}

\subsection{Performance on Neuromorphic Datasets}

\begin{figure}[h]
  \centering
  \includegraphics[width=0.95 \textwidth]{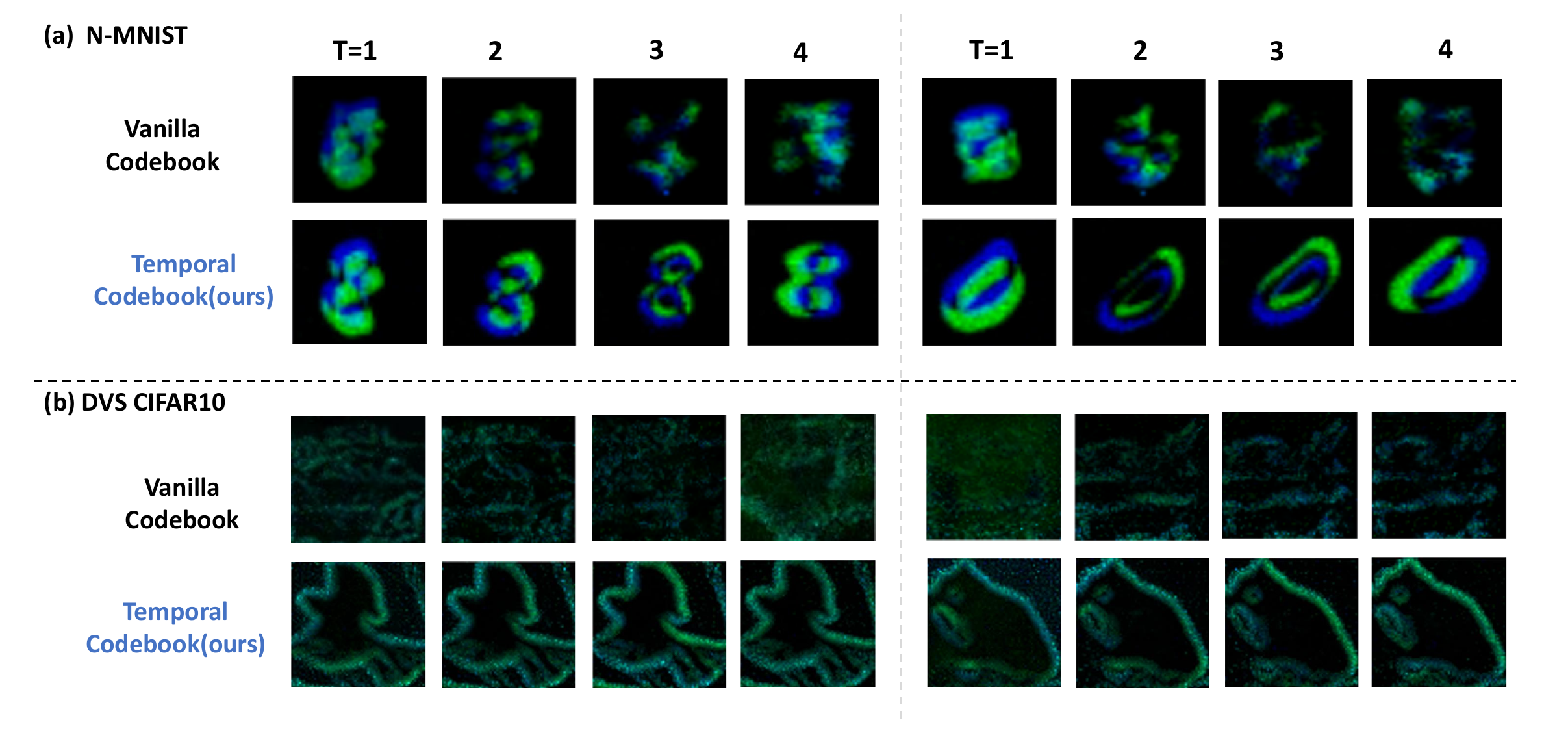}
  \caption{Temporal unfolding and generation quality comparison on neuromorphic datasets using temporal and vanilla codebooks on N-MNIST (a) and DVS-CIFAR10 (b). }
  \label{fig:dvs}
\end{figure}

To validate the effectiveness of our proposed method on neuromorphic datasets, we conducted experiments on the N-MNIST and DVS-CIFAR10 datasets. The results are illustrated in Figure~\ref{fig:dvs}.
Figure~\ref{fig:dvs}(a) displays the temporal unfolding of the generated results on NMNIST~\cite{orchard2015converting}, while Figure~\ref{fig:dvs}(b) shows the results on DVS-CIFAR10~\cite{li2017cifar10}. It is evident that, compared to the vanilla codebook, the temporal codebook yields superior results both in terms of the quality of generation at individual time steps and in reducing temporal inconsistency.

\subsection{High-Resolution Image Generation}

\begin{figure}[h]
  \centering
  \includegraphics[width=0.95\textwidth]{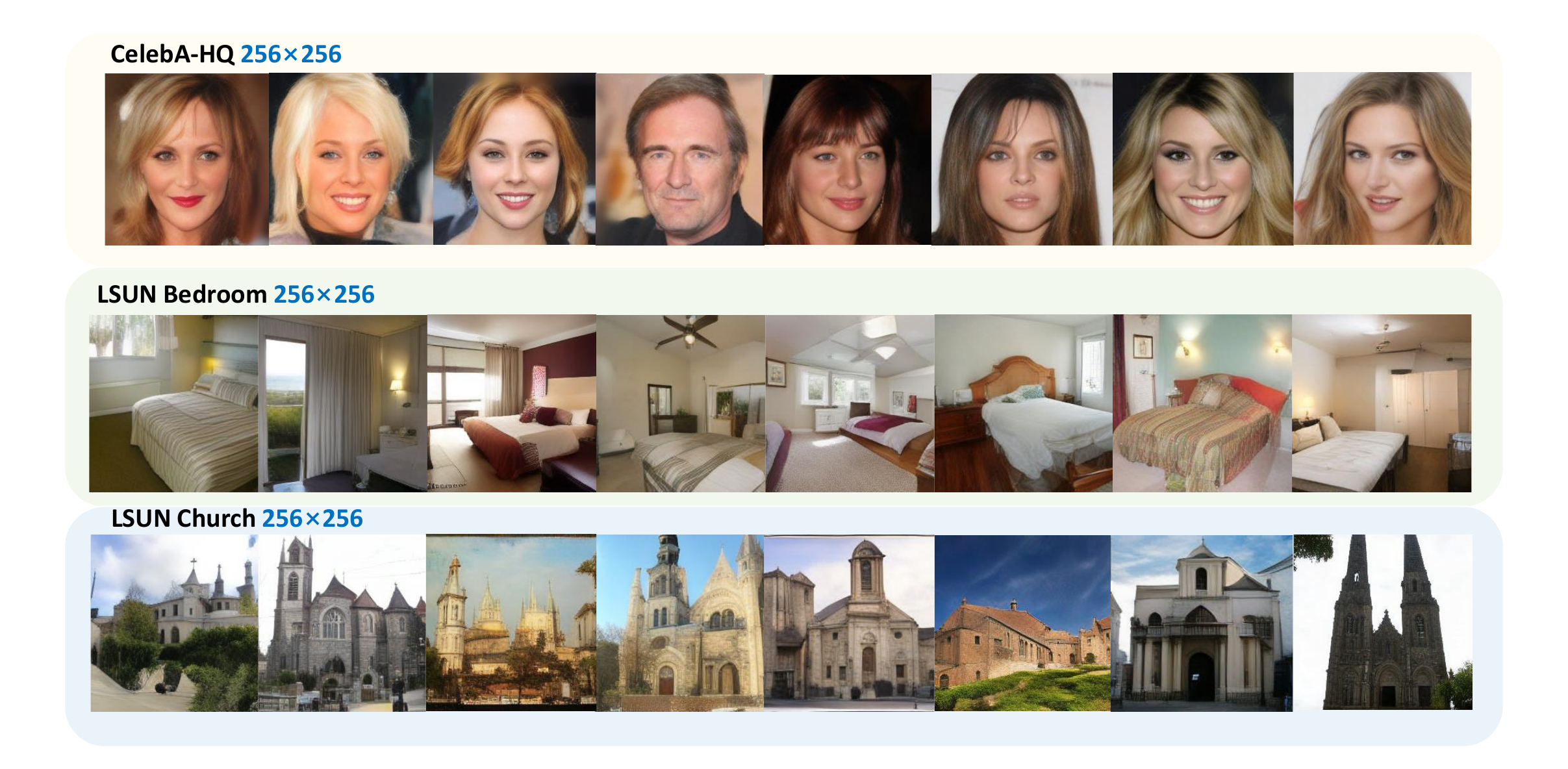}
  \caption{Generated high-resolution images using the proposed SNN-based method}
  \label{fig:show_img}
\end{figure}

To validate the performance of our method on high-resolution images, we conducted experiments on higher-resolution datasets, specifically the \(256 \times 256\) CelebA-HQ~\cite{karras2017progressive}, LSUN Bedroom, and LSUN Church~\cite{yu15lsun} datasets. Our study is the first to successfully employ SNNs for generating high-resolution images on the aforementioned datasets. We assessed the generated images using various metrics, including FID, Precision and Recall, and Kernel Inception Distance (KID), to provide a comprehensive evaluation and facilitate further research. The quantitative results are summarized in Table~\ref{tab:res}. The selections of generated samples are shown in Figure~\ref{fig:show_img}. 

% \subsection{Results on High-Resolution Images}
% Previous generative models based on SNNs~\cite{kamata2022fully,feng2023sgad,liu2023spiking} are limited to producing outputs on lower-resolution datasets, such as the \(28 \times 28\) MNIST dataset, the \(32 \times 32\) CIFAR-10 dataset, or the \(64 \times 64\) CelebA~\cite{liu2018large} dataset. In our study, we have for the first time successfully utilized SNNs to generate data on higher-resolution datasets, specifically the \(256 \times 256\) CelebA-HQ~\cite{karras2017progressive}, LSUN Bedroom, and LSUN Church~\cite{yu15lsun} datasets. We evaluate the generated results using a suite of metrics including Fréchet Inception Distance, Precision and Recall, and Kernel Inception Distance, to facilitate further research building on our findings. The quantitative results are presented in Table~\ref{tab:res}. In Figure~\ref{fig:show_img}, we display some of the generated images. To further compare our results with other works, we also conducted experiments on the \(64 \times 64\) CelebA and LSUN Bedroom datasets.

% \begin{table}[H]
%   \centering
%   \begin{tabular}{@{}cccccc@{}}
%     \toprule
%     \textbf{Resolution}           & \textbf{Dataset}      & \textbf{FID} $\downarrow$ & \textbf{Precision} $\uparrow$ & \textbf{Recall} $\uparrow$ & \textbf{KID} $\downarrow$ ($\times 10^{-3}$) \\ \midrule
%     \multirow{3}{*}{\textbf{256}} & CelebA-HQ    & 12.98           & 0.770               & 0.410            & 7.78                               \\ \cmidrule(l){2-6}
%                          & LSUN Bedroom & 6.61            & 0.630               & 0.405            & 3.98                               \\ \cmidrule(l){2-6}
%                          & LSUN Church  & 6.62            & 0.675               & 0.465            & 3.22                               \\ \midrule
%     \multirow{2}{*}{\textbf{64}}  & CelebA       & 13.57           & 0.884               & 0.182            & 10.26                              \\
%                          & LSUN Bedroom & 12.90           & 0.795               & 0.141            & 8.83                               \\ \bottomrule
%   \end{tabular}
%   \caption{Quantitative evaluation metrics for high-resolution image generation}
%   \label{tab:res}
% \end{table}

\begin{table}[h]
\centering
\begin{tabular}{@{}cccccc@{}}
\toprule
\textbf{Resolution}           & \textbf{Dataset} & \textbf{FID} $\downarrow$ & \textbf{Precision} $\uparrow$ & \textbf{Recall} $\uparrow$ & \textbf{KID} $\downarrow$ ($\times 10^{-3}$) \\ \midrule
\multirow{3}{*}{\textbf{256}} & CelebA-HQ        & 12.98                     & 0.770                         & 0.410                      & 7.78                                         \\ \cmidrule(l){2-6} 
                              & LSUN Bedroom     & 6.61                      & 0.630                         & 0.405                      & 3.98                                         \\ \cmidrule(l){2-6} 
                              & LSUN Church      & 6.62                      & 0.675                         & 0.465                      & 3.22                                         \\ \bottomrule
\end{tabular}
\caption{Quantitative evaluation metrics for high-resolution image generation}
\label{tab:res}
\end{table}

\subsection{Effectiveness of the Temporal Codebook}
In this section, we empirically demonstrate the significant impact of the temporal codebook on generation results. We first varied the time steps and measured the FID for two datasets, CelebA and LSUN Bedroom ($64 \times 64$), employing both the temporal codebook and a vanilla codebook. The results are presented in Table~\ref{tab:te}.

\begin{table}[H]

  \centering
  \begin{tabular}{@{}ccccc@{}}
    \toprule
    \multirow{2}{*}{\textbf{Dataset}}                                                                   & \multirow{2}{*}{\textbf{Codebook}} & \multicolumn{3}{c}{\textbf{FID} $\downarrow$}                                   \\ \cmidrule(l){3-5}
                                                                                               &                           & T=2                                 & T=4            & T=6            \\ \midrule
    \multirow{2}{*}{CelebA}                                                                    & Vanilla                   & 28.00                               & 67.39          & 73.09          \\
                                                                                               & \textbf{Temporal}                  & 14.87                               & \textbf{13.57} & 13.69          \\ \midrule
    \multirow{2}{*}{\begin{tabular}[c]{@{}c@{}}LSUN Bedroom\\ (\(64 \times 64\))\end{tabular}} & Vanilla                   & 44.04                               & 105.66         & 121.38         \\
                                                                                               & \textbf{Temporal}                  & 15.24                               & 14.74          & \textbf{12.90} \\ \bottomrule
  \end{tabular}
  \caption{Comparison of the FID between the temporal codebook and the vanilla codebook on the \(64 \times 64\) CelebA and LSUN Bedroom datasets.}
  \label{tab:te}
\end{table}

The results reveal that the temporal codebook, enriched with temporal information, significantly enhances model performance compared to the vanilla codebook, highlighting the critical importance of temporal information in SNN-based generative models. Moreover, it is evident that increasing the time steps with the temporal codebook improves the FID, an effect not observed with the vanilla codebook. This further substantiates that the temporal codebook effectively integrates temporal information, and extending the time steps enables the model to generate images more accurately.

\begin{figure}[htbp!]
  \centering
  \includegraphics[width=0.8 \textwidth]{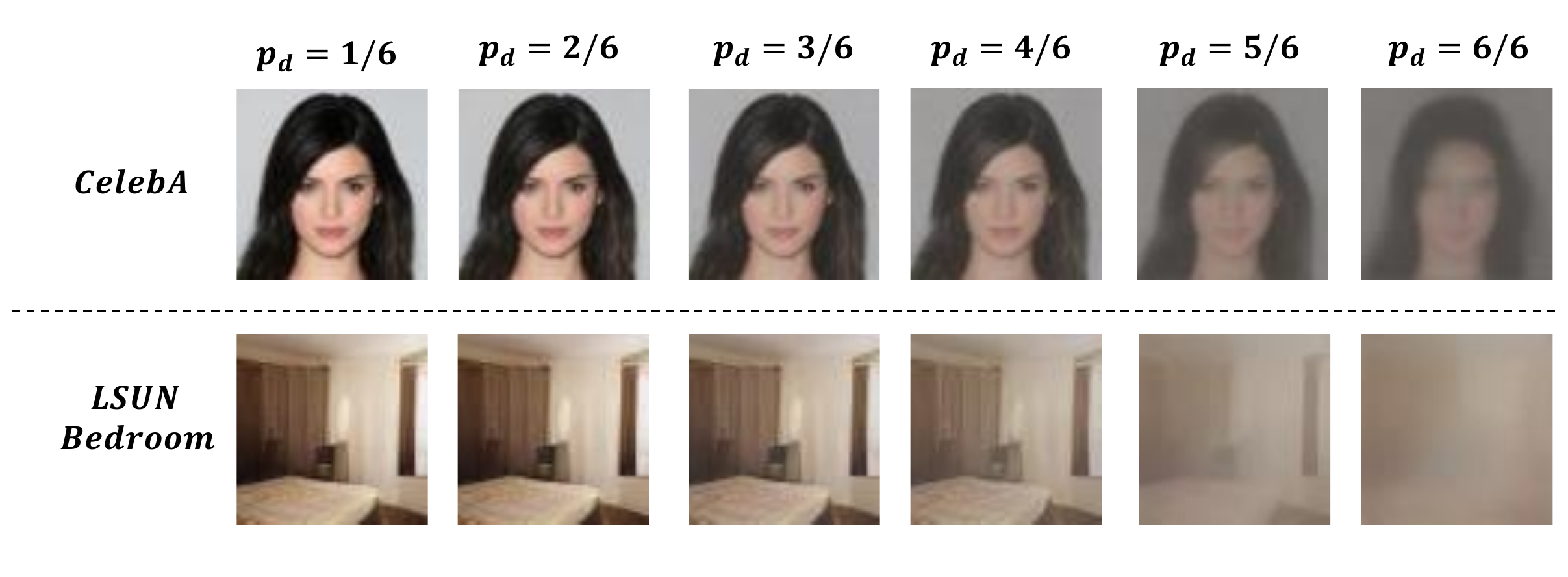}
  \caption{Effect of temporal destruction on image quality using temporal codebook. Visualizations for various values of \( p_d \), the temporal destruction factor, show minimal impact on image quality at lower \( p_d \) values, with significant deterioration observed at higher \( p_d \) values.}
  \label{fig:destruction}
\end{figure}

To further validate the effectiveness of the temporal codebook, we conducted a destructive experiment wherein vectors within the temporal codebook were randomly substituted with information from the last temporal steps. A temporal destruction factor \( p_d \) was introduced to define the proportion of temporal information to be replaced with alternate temporal data. Visualizations for various values of \( p_d \) are depicted in Figure~\ref{fig:destruction}. If \(p_d = \frac{2}{6}\), it implies that out of a total of 6 time steps, 2 time steps are replaced with information from other time steps. The results indicate that at lower values of \( p_d \), such as \( p_d = \frac{1}{6} \) and \( p_d = \frac{2}{6} \), the impact on image quality is minimal. However, as \( p_d \) increases, there is a significant deterioration in image quality, particularly between \( p_d = \frac{4}{6} \) and \( p_d = \frac{5}{6} \).

% \begin{table}[H]
%   \centering
%   \begin{tabular}{@{}ccccccc@{}}
%     \toprule
%     \multirow{2}{*}{\textbf{Dataset}} & \multicolumn{6}{c}{\textbf{FID} $\downarrow$}                                                             \\ \cmidrule(l){2-7}
%                              & $p_d=1/6$                           & $p_d=2/6$ & $p_d=3/6$ & $p_d=4/6$ & $p_d=5/6$ & $p_d=6/6$ \\ \midrule
%     CelebA                   & 13.69                               & 13.04     & 14.43     & 18.40     & 47.53     & 143.10    \\ \midrule
%     LSUN bedroom             & 12.90                               & 13.05     & 13.99     & 15.79     & 95.25     & 351.23    \\ \bottomrule
%   \end{tabular}
%   \caption{By replacing the temporal information in vectors from the temporal codebook with information from other temporal positions at a proportion defined by \( p_d \), the relationship between the FID and the variable \( p_d \) was observed.}
%   \label{tab:destruction}
% \end{table}

\begin{figure}[h]
  \centering
  \includegraphics[width=0.75 \textwidth]{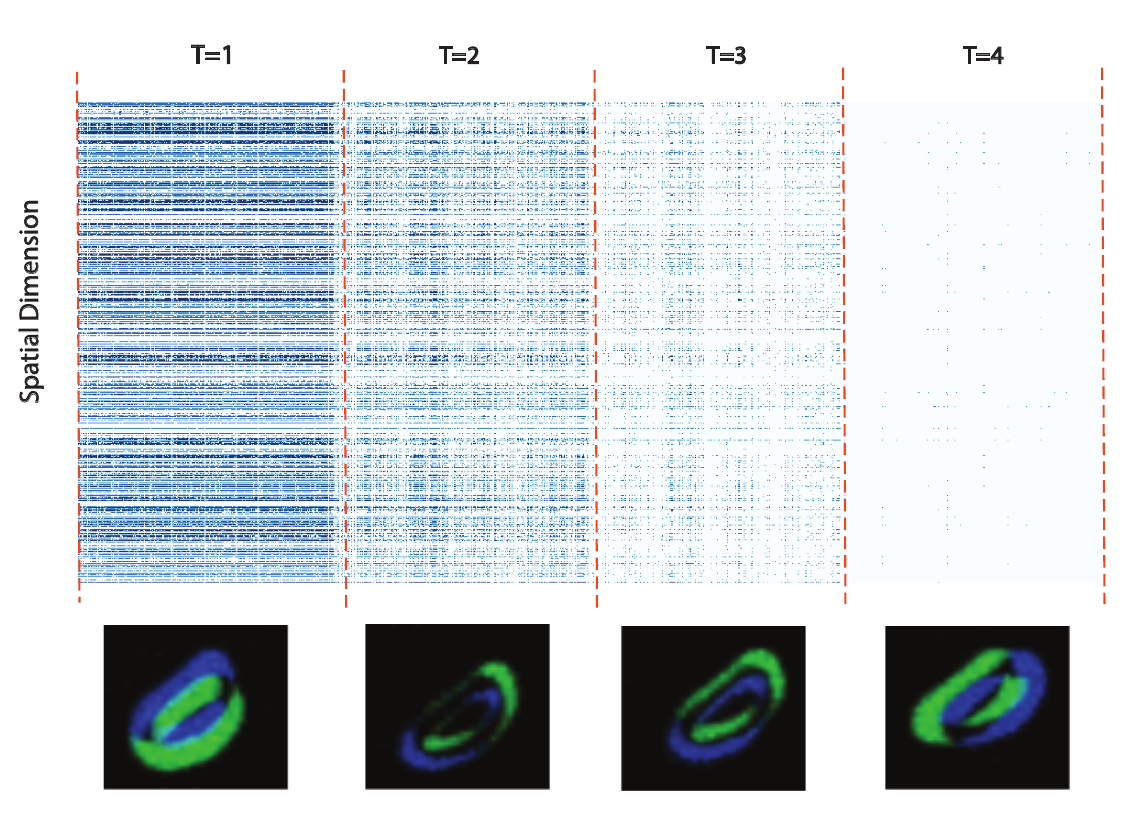}
  \caption{Visualization of the magnitude of elements in the temporal codebook \(\tilde{Q}\)}
  \label{fig:decay}
\end{figure}

We visualized the temporal codebook \(\tilde{Q}\) by plotting a heatmap based on the magnitude of the elements in \(\tilde{Q}\), as shown in Figure~\ref{fig:decay}. The figure illustrates that as the time steps increase, the activity level of the corresponding elements in \(\tilde{Q}\) decreases. This decreasing activity mirrors the behavior of biological time cells, which exhibit diminishing activity over a certain time interval after receiving a stimulus.

\section{Conclusion}
In this work, we introduced a novel SNN-based VQ-VAE framework that incorporates a temporal codebook inspired by hippocampal time cells. Our extensive experiments demonstrate the superior performance of our approach in generating high-quality, temporally consistent images across various datasets. Specifically, our method achieved state-of-the-art results on high-resolution datasets and showed significant improvements in temporal correlation and image quality compared to existing SNN-based generative models. We also validated the efficacy of the temporal codebook through destructive experiments, which underscored its crucial role in leveraging temporal information to enhance model performance. Our findings suggest that integrating temporal dynamics into the latent space of generative models holds great potential for advancing the field of neuromorphic computing and spiking neural networks. Future work will explore further optimizations and applications of this framework to other temporally rich generative tasks. However, there are some limitations to our approach. The complexity of the temporal codebook may increase computational burden. Future work will focus on optimizing this framework and exploring its applications to other temporally rich generative tasks.

\newpage
\bibliographystyle{plain}
\bibliography{myref}

\end{document}